\documentclass{article}

\usepackage[square,numbers]{natbib}
\usepackage[final]{./nips_2017_nofooter}

\usepackage[utf8]{inputenc} 
\usepackage[T1]{fontenc}    
\usepackage{hyperref}       
\usepackage{url}            
\usepackage{booktabs}       
\usepackage{amsfonts}       
\usepackage{nicefrac}       
\usepackage{microtype}      

\usepackage[pdftex]{graphicx}
\usepackage{subcaption}
\usepackage{algorithm}
\usepackage[noend]{algpseudocode}
\usepackage{color}

\title{Learning human behaviors from motion capture\\ by adversarial imitation}

\author{
  Josh Merel, Yuval Tassa, Dhruva TB, Sriram Srinivasan, Jay Lemmon, Ziyu Wang,\\
  \textbf{Greg Wayne, Nicolas Heess} \\
  DeepMind\\
  \texttt{jsmerel,tassa,dhruvat,srsrinivasan,numsgil,ziyu,gregwayne,heess@google.com}
}

\begin{document}

\maketitle

\begin{abstract}

Rapid progress in deep reinforcement learning has made it increasingly feasible to train controllers for high-dimensional humanoid bodies. 
However, methods that use pure reinforcement learning with simple reward functions tend to produce non-humanlike and overly stereotyped movement behaviors.  In this work, we extend generative adversarial imitation learning to enable training of generic neural network policies to produce humanlike movement patterns from limited demonstrations consisting only of partially observed state features, without access to actions, even when the demonstrations come from a body with different and unknown physical parameters.  We leverage this approach to build sub-skill policies from motion capture data and show that they can be reused to solve tasks when controlled by a higher level controller.
[\href{https://youtu.be/YsxN3uRBupc}{{\color{blue}video abstract}}]
\end{abstract}

\section{Introduction}

The problem of building a programmable humanoid dates back centuries. In 1495, five years after drawing the Vitruvian Man, Leonardo da Vinci constructed a humanoid automaton in the form of an armored knight \cite{rosheim2006leonardo}. The knight was able to wave, sit up, and open and close its jaw via power delivered by a crank. Unlike most clockwork automata, which could only produce movements along individual limit cycles, the mechanical knight could be re-programmed to vary its movements, enabling refinement of the arm movements or alternative sequences of movements in time. 

From a contemporary perspective, optimal control and reinforcement learning methods are enabling the design of movement controllers that can cope with the high-dimensionality of humanoid bodies, and neural networks are able to store multiple patterns of movement that can be reused, refined, and flexibly sequenced. Working towards a robust procedure for constructing controllers with a range of humanlike movements suited for reuse and refinement when employed in new tasks is the goal of this paper.

Yet current methods for humanoid control fall short of our desires. Methods that rely on pure reinforcement learning (RL) objectives tend to produce insufficiently humanlike and overly stereotyped movement behaviors.
The uncanniness of these movements can be improved with meticulous body design (e.g. muscles), controller specialization (e.g. phase variables), and reward function engineering, but these methods require substantial domain expertise.
Designing reward functions to capture the intricacies of humanoid behavior is difficult and must be repeated for every new behavior.

A compelling alternative to RL, commonly employed in the computer animation literature, is to use motion capture data as demonstrations to load movements into controllers. 
These methods typically produce visually pleasing, humanlike movements.
However, existing methods for using motion capture come with caveats. Some of these methods use kinematic sequence models, and merely produce sequences of states (possibly violating physical constraints); some methods only produce tracking controllers for specific trajectories; other methods require heavily engineered components like cost functions that dictate how much to weight matching different features of the movement trajectory; yet others reduce model flexibility by designing the parameterizations of gaits or body-environment interaction.

In order to perform imitation learning from motion capture data, we make heavy use of generative adversarial imitation learning (GAIL; \cite{ho2016generative}), which is a recent breakthrough in imitation learning that produces an imitation policy in a manner similar to generative adversarial networks (\cite{goodfellow2014generative}). State-action pairs from demonstration data are compared against state-action pairs from the policy. A classifier is trained to discriminate demonstration data from the imitation data, and the imitation policy is given reward for fooling the discriminator.
The key benefit of GAIL over previously existing imitation-learning techniques is that the notion of similarity between imitation and demonstration data does not have to be defined based on an explicit, hand-designed metric.

\begin{figure}[t!]
  \centering
  \includegraphics[width=1\linewidth]{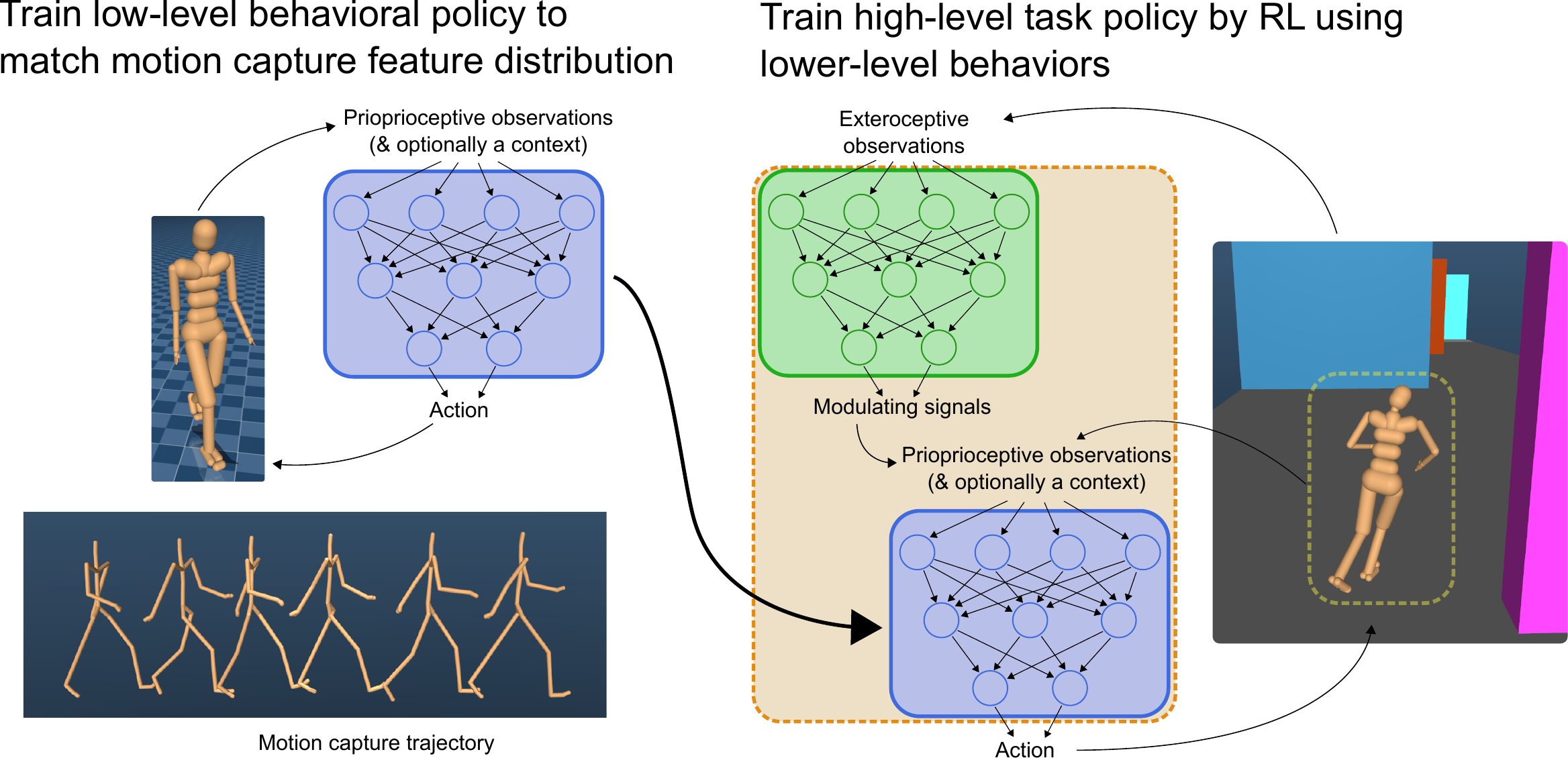}
  \caption{Overview of our approach: (Left) First train specific skills into low-level controller (LLC) policies by imitation learning from motion capture data. (Right) Train a high-level controller (HLC) by RL to reuse pre-trained LLCs.}
  \label{overview}
\end{figure}

In this work, we present a pipeline for (1) training low-level controllers to produce behaviors from motion capture using an extension of GAIL; and (2) embedding the low-level controllers into larger control systems wherein a high-level controller learns by RL to modulate the low-level controller to solve new tasks (Figure \ref{overview}).  
The acquisition of multiple behaviors from noisy motion capture data (``real-to-sim") requires two extensions to the GAIL framework.  The original presentation of GAIL was restricted to imitation of single skills from complete state-action trajectories, where the demonstrator shared the same body and policy parameterization as the imitator. We demonstrate: (a) partial state featurizations without demonstrator actions suffice for adversarial imitation; (b) the body structure and physical parameters (i.e. body dynamics) need not match between the demonstrator and the imitator; and (c) robust transitions between behaviors naturally emerge by training on multiple behaviors.  

Taken together, our approach produces reusable humanlike skills extracted from very small quantities of noisy human motion capture demonstrations without a great deal of domain engineering.

\section{Methods}

\begin{figure}[t!]
  \centering
  \includegraphics[width=1\linewidth]{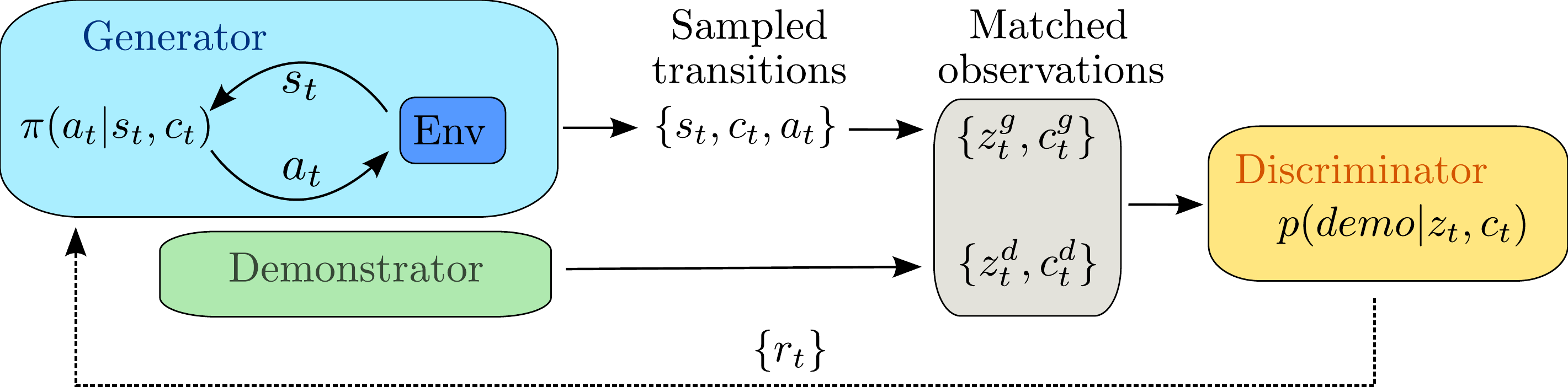}
  \caption{GAIL framework with the addition of context variable, $c$, for multi-behavior policies. A stochastic policy $\pi$ interacting with an environment produces trajectories of states and actions (analogous to generator in GAN framework).  The state-action pairs are transformed into features, $z$, which we show may exclude actions.  The demonstration data are assumed to be in the same feature space.  Either demonstration data or generated data are evaluated by the discriminator to yield a probability of the data being demonstration data. The discriminator provides a reward function for the policy.}
  \label{diagram}
\end{figure}

\textbf{Adversarial imitation learning:} \citet{ho2016generative} show that their approach, GAIL, encourages the imitator to match the state-occupancy distribution of the demonstrator. In that work, demonstration data were generated by training a first policy via RL on a task with a hand-designed reward function, logging trajectories, and training a second policy of the same architecture as the first to imitate it by GAIL. While impressive, those results constitute a validation of the algorithm in the most favorable setting -- the same body, simulator, and policy architecture are used to produce demonstrations as are used for imitation. In this paper, we demonstrate that adversarial imitation learning can work even when the discriminator only has access to states (not actions) as well as partial state observations. Furthermore, for observed features that are sufficiently invariant across bodies, we do not need to know the underlying physical parameters or dynamics of the body that gave rise to the observations, allowing imitation in more general contexts.

We describe our variant of generative adversarial imitation learning (GAIL, \cite{ho2016generative}). We extend GAIL to settings where we have only partial observations of state or features thereof, excluding actions, and provide a context label to the discriminator to allow for training of multiple behaviors simultaneously (which allows for robust transitioning between behaviors, see Results). We present the architecture diagrammatically in Figure \ref{diagram} and the algorithm as pseudocode (Algorithm \ref{algorithm}). 

We parameterize a stochastic policy with a neural network which generates a Gaussian distribution for each actuator, from which actions are sampled. Both the mean and log-standard-deviation of the action model are trainable.  To train this policy, we use the adversarial imitation learning algorithm. Each iteration consists of three essential steps: (1) collect data using the current policy; (2) update the policy using rewards determined by the discriminator; (3) update the discriminator using the demonstration data and generated data. As in \cite{ho2016generative}, policy updating is performed by trust-region policy optimization (TRPO; \cite{schulman2015trust}). TRPO is particularly convenient in this case due to robustness and its need for only limited hyperparameter tuning, which allows hyperparameter searches to focus on GAN-related hyperparameters (e.g. the relative rate of discriminator updating).

In contrast to the common practice for GAN training, we found that updating the policy with a reward proportional to $-\log(1-D(\cdot))$ worked better than a reward proportional to $\log(D(\cdot))$. Other details of policy training are provided in the supplement. 

\begin{algorithm}[h!]
\caption{}
\label{algorithm}
\begin{algorithmic}
    \State \textbf{Input:} Set of demonstration observations $\{z_t^d,c_t^d\}_{t=1\dots T^d }$
    \State Randomly initialize policy ($\pi_{\theta}$) and discriminator ($D_{\phi}$)
    \State // Perform N training iterations of policy \& discriminator updating
    \For {$i$ in $1\dots N$}
    \State Execute policy rollouts to collect $T^g$ timestep observations, $\{z^g_t,c^g_t\}_{t= 1\dots T^g}$
    \State Compute rewards $\{r_t = -\log(1-D_{\phi}(z^g_t,c^g_t))\}_{t=1\dots T^g}$
    \State Update $\theta$ (e.g. by TRPO)
    \State // Perform M discriminator updates steps
    \For {$j$ in $1\dots M$}
        \State $\ell(\phi) = \sum_{t=1 \dots T^g} \log(1-D_\phi(z_t^g,c_t^g)) - \sum_{t=1 \dots T^d} \log(D_\phi(z_t^d,c_t^d))$
        \State Update $\phi$ by a gradient method w.r.t. $\ell(\phi)$  
        \EndFor
    \EndFor
	\State \textbf{Return:} $\pi$
\end{algorithmic}
\end{algorithm}

\textbf{Simulation:} All simulations make use of the physics engine MuJoCo (\cite{todorov2012mujoco}).  In this paper, we consider four \emph{bodies}: a bipedal walker, restricted to lateral movement in a vertical plane; two-link and three-link planar arms; and a custom humanoid body introduced below. 
Bodies consist of mass-bearing segments (\emph{geoms}), which are connected by \emph{joints}.  
The \emph{root} of a body translates and rotates freely and is not actuated.  All other joints are torque-actuated, and it is the role of the \emph{policy} (or controller) to determine what torques to generate.  

\textbf{Motion capture demonstrations:} The CMU Graphics Lab Motion Capture Database \href{http://mocap.cs.cmu.edu/}{{\color{blue}(link)}} provides motion capture models and data for multiple subjects and a wide range of behaviors.
We programmatically generated a humanoid body (56 joint-angle DOFs, and a 6 DOF root joint) based on one of the subjects.
Although we iteratively and interactively evaluated the general physical plausibility of the body we designed, there is almost surely only limited resemblance between the dynamical properties of the body we produced and those of the actual human whose body produced the motion capture data.

We use motion capture data from arbitrary subjects, which means we are re-targeting motion capture data across subjects (i.e. mapping joint angles from one body to another).  After very limited preprocessing, the demonstration trajectories are dynamically inconsistent and include clear physical violations (penetration and/or floating), so any approach that makes use of this data must be robust to these issues.  All experiments use whole clips from the CMU database, with no additional segmentation.

\section{Results}
To accomplish imitation learning from human motion capture, we need to show that GAIL can work even when imitator and demonstrator bodies differ and raw control actions are unknown.
We first present results using simpler environments to demonstrate that GAIL can be extended to these settings.
For these validation experiments, we first train policies via RL using hand-designed reward functions to solve simple tasks and then train imitation policies based on logged demonstrations.  
This strategy helps us to quantify performance because while we lack objective metrics for assessing the GAIL-trained imitation policies, we can assess imitation policies using the task objective originally used to train the demonstration policy.
We then construct policy sub-skills for the complex humanoid using motion capture and then explore the reuse of the sub-skills in the context of tasks inside environments.

\subsection{Validation of imitation of without actions, using partial observations}

\begin{figure}[t!]
  \centering
  \includegraphics[width=0.49\linewidth]{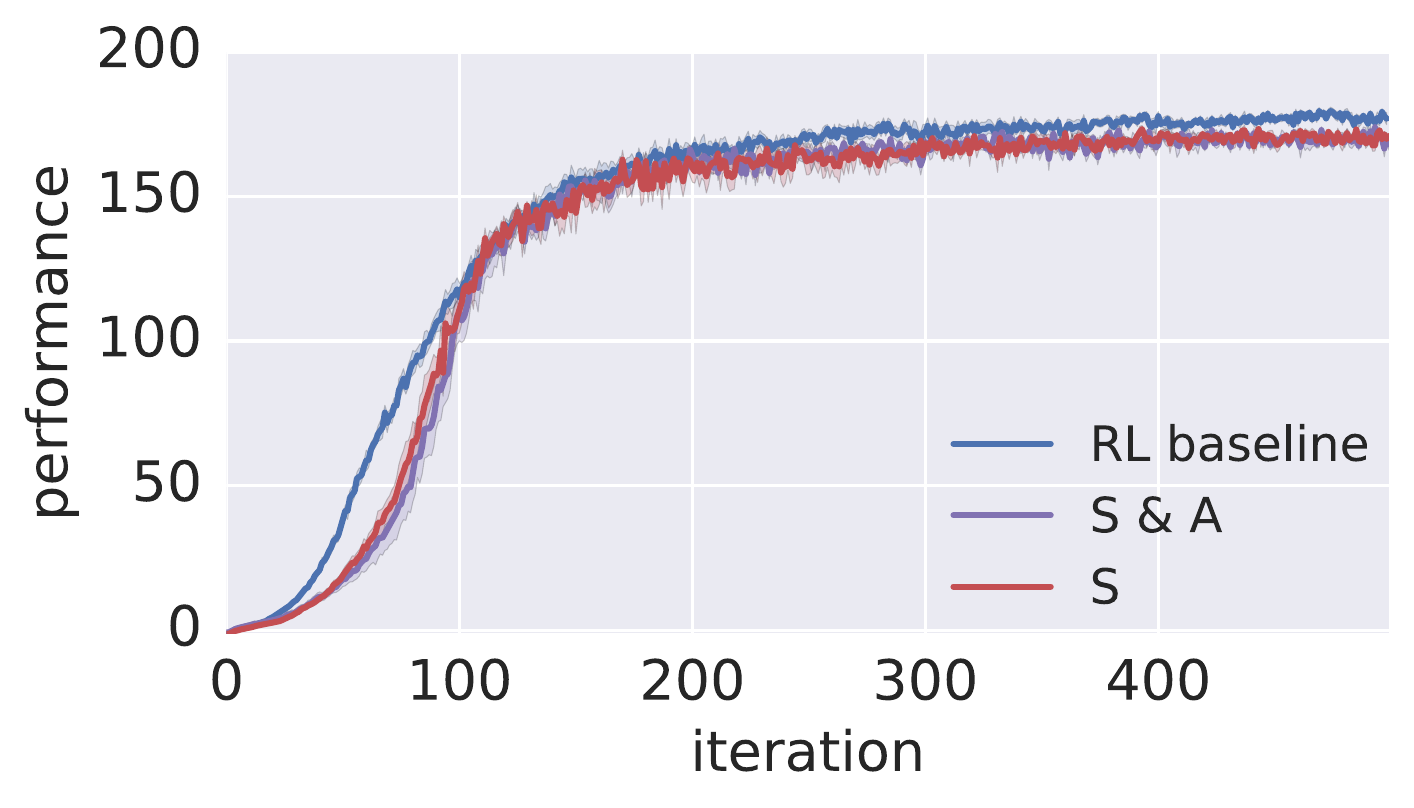}
  \includegraphics[width=0.49\linewidth]{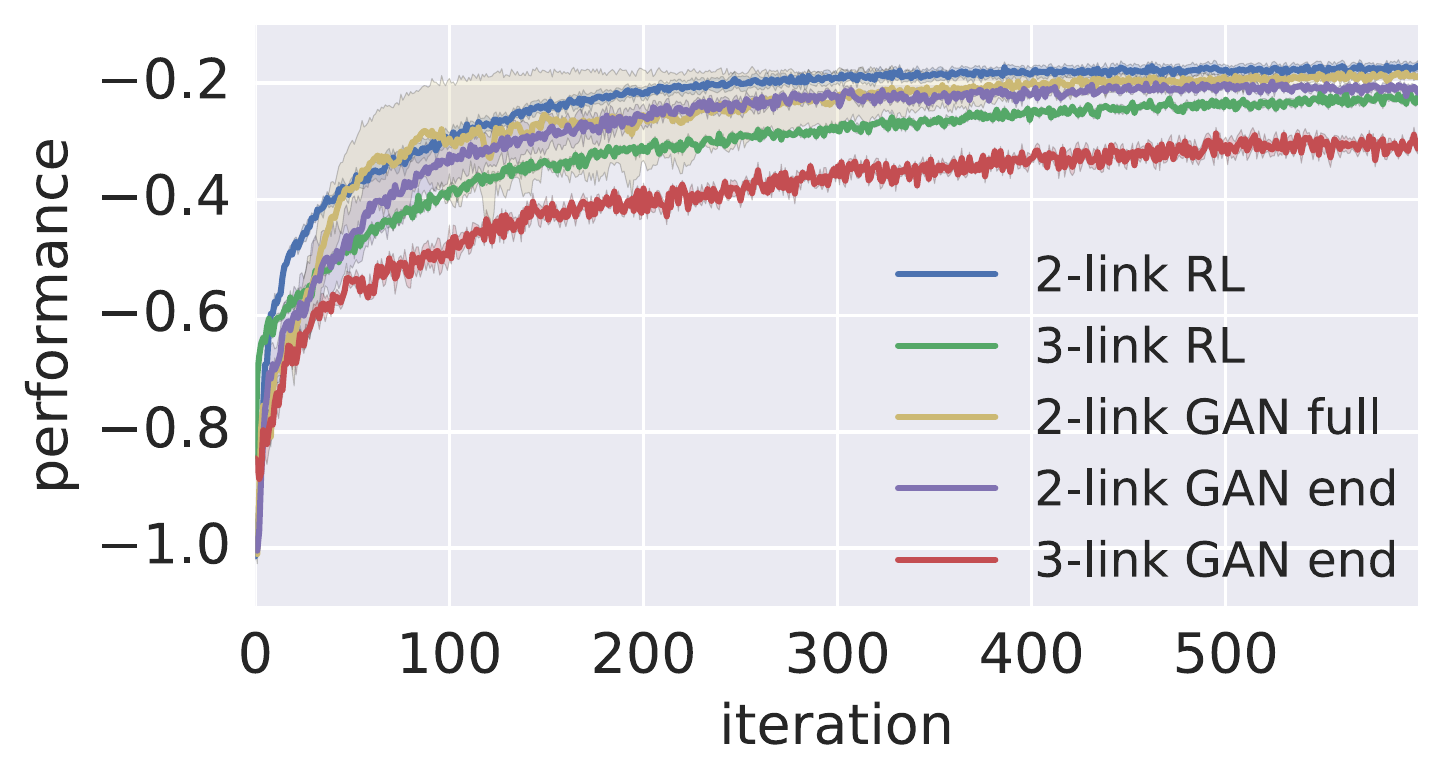}
  \caption{(Left) Comparison of walker trained by RL versus various imitation-trained policies with actions (i.e. states and actions, S \& A) or without actions (states only, S).  For tuned  hyperparameters (matched), the imitator policies perform well regardless of action access during training.  
  (Right) Comparison of two and three link arms trained by RL versus imitation-trained policies using full or partial state access.  Policies are successfully trained with partial states and across bodies.
  All curves are averages of 5 seeds.}
  \label{basic_validation}
\end{figure}

\textbf{Imitation without ground-truth actions:} To show that it is sufficient for the discriminator to condition on state information (including velocity) without accompanying actions, we examine imitation learning for a 2D planar walker (see Figure \ref{basic_validation}, left panel).
The walker task consists of 10s episodes, terminating early if the walker torso falls below a threshold. The demonstration policy is continuously rewarded proportionally to the absolute difference between its horizontal velocity and a target speed (5m/s), minus a small control cost. See \href{https://www.youtube.com/watch?v=73pCwtZAA9A}{{\color{blue}video}} of the demonstration, and see supplemental information for additional details.
We observe that comparing actions in addition to states by the discriminator provides no benefit for training the imitation policy. 

To interpret this result, consider that for fully-actuated bodies in deterministic environments, there exists an inverse mapping between two successive states and the action connecting them. In such a case, the action can be inferred directly from changes of state. More generally, together with the constraints imposed by the form of the body and environment, even without full-actuation or determinism, reasonable actions can be inferred.
Furthermore, when the imitator and demonstrator bodies do not match, naive use of action information would actually be detrimental for imitation as the actions may no longer produce similar physical effects.

\textbf{Features invariant to body transfer:} Having dispensed with actions, we are also interested in understanding the extent to which imitation can be performed across bodies, which is sometimes referred to as ``re-targeting''. 
As human observers, we expect certain correspondences between equivalent behaviors across bodies.  
For example, a human may wave to a dolphin and would be gratified to see it ``wave'' back with a fin, despite different numbers of actuators/DOFs between the respective bodies and ``end-effectors''.  Here, we simply examine whether state-only demonstrations of the position of a two-link arm end-effector sufficiently constrain the re-targeted imitation for an arm with three-links (Figure \ref{basic_validation}, right panel). 

The task consists of 4s episodes during which the target position changes every second. 
The reward at each timestep is proportional to squared distance between the end-effector and the target center minus a small control cost.  We provide videos depicting performance of RL baselines on \href{https://www.youtube.com/watch?v=H7ICmKTD_TM}{{\color{blue}two-link}} and \href{https://www.youtube.com/watch?v=6j58KSRr9rg}{{\color{blue}three-link}} arm tasks. See supplemental information for additional details.
We again log demonstrations, this time from the RL-trained two-link arm, and we train imitator policies which control either the two- or three-link arms.    
For imitation, the policy still receives full observation, but the discriminator only has access to a subset of the observation. We see that the two-link arm is able to imitate demonstrations from another two-link arm roughly equivalently well using either the full set of observations or observing only the vector from end-effector to target center (Figure \ref{basic_validation}, right panel, yellow vs. purple curves). 
The three-link arm can also be trained to imitate the two-link arm if the discriminator observes only the feature representing the displacement from the end-effector to target vector (Figure \ref{basic_validation}, right panel, red curve, three-link imitation \href{https://www.youtube.com/watch?v=iycWkfgL-D0}{{\color{blue}video}}). 
The choice of the vector from the end-effector to the target as a feature corresponds to the selection of a feature that is nearly invariant to body structure. 
Thus, the imitator can learn to make the three-link arm move such that it causes the vector between end-effector and target target to match the statistics of the demonstrations.

\subsection{Training a complex humanoid from motion capture}

\begin{figure}[t!]
  \centering
  \includegraphics[width=1\linewidth]{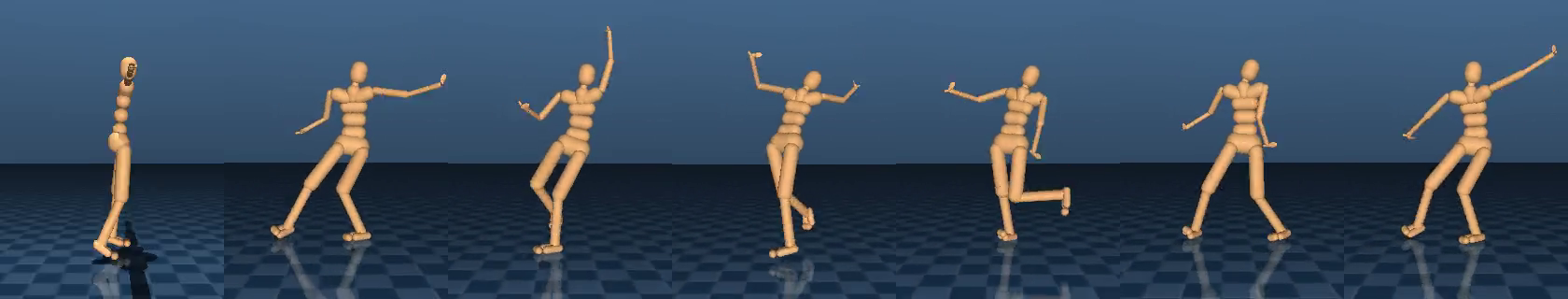}
  \includegraphics[width=1\linewidth]{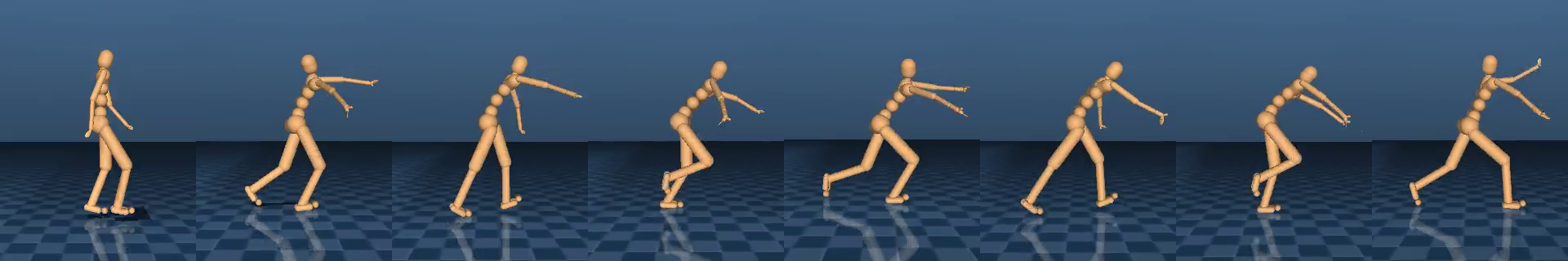}
  \caption{Comparison of RL policies trained to move forwards (i.e. rightwards on the page) from (top) randomized versus (bottom) motion capture pose initialization at the beginning of episodes during training.
  Leftmost tile per row is the initial pose in probe episode followed by equispaced frames of a representative gait during that episode. 
  Note the extremely unnatural sideways skip depicted in the top panel versus the still awkward, but more natural gait depicted in the bottom panel.
}
  \label{initializations}
\end{figure}

Having validated extensions to the adversarial imitation learning approach on relatively simple problems, we are now in a position to train policies from motion capture data.  
We emphasize that, compared with other recent work optimizing generic policies for humanoid bodies by RL (e.g. \cite{heess2016learning, schulman2015high}), the presently studied body is quite complex, with more than twice as many action dimensions (56 actuated dimensions, vs. 21). 
See supplemental information for humanoid experimental procedures.

\textbf{Episode initialization:} We must first determine the distribution of states for the beginning of episodes. Naively, the initial state distribution could be manually engineered to be some specific starting pose with small variability.  When training using RL, we observe that unnatural initial poses (whether \href{https://www.youtube.com/watch?v=hrVAK8c8bFM}{{\color{blue}fixed}} or \href{https://www.youtube.com/watch?v=CNtBFf8MToQ}{{\color{blue}variable}}) can yield different, unnatural locomotion behavior. In our setting where we expect to learn from motion capture, it is reasonable to initialize to poses randomly sampled from motion capture data. If we merely initialize the body from the motion capture poses and train from an RL objective to run forward, we can already observe a visually striking difference in the naturalness of the gait, though the gait is still fairly non-humanlike (see Fig \ref{initializations}).  (see \href{https://www.youtube.com/watch?v=3RTOSMAeBas}{{\color{blue}video}}). This effect is both dramatic and reliable. 

\textbf{Imitation from partial observations:} Our first attempt at imitation learning made use of a small set of motion capture data consisting of clips of a few steps of walking (<30s human data; supplement for details).
Observations available to the policy initially consisted of non-root joint angles and joint velocities as well as absolute velocity (3D root velocity) -- this constitutes a full observation of the body/environment and had worked fine for RL.  
However, naive application of adversarial imitation, where the discriminator receives all features available to policy, did not work well (see \href{https://www.youtube.com/watch?v=1f4Ycsm7th8}{{\color{blue}video}} for representative results of diverse policies unsatisfactorily trained by adversarial imitation from walking behavior). 
This failure can be traced to the fact that the motion capture demonstrations have noise and are dynamically inconsistent for the imitator body. We observe in particular that the arms (and the legs to a lesser extent) move in an unnaturally jerky fashion, suggesting that discriminator comparison based on joint angles poorly constrains the behavior, perhaps due to compounding errors in the sequence of joint angles from the torso to the hands.  

\begin{figure}[!t]
    \begin{tabular}[t]{cc}
\begin{subfigure}{0.2\textwidth}
    \includegraphics[width=1\linewidth]{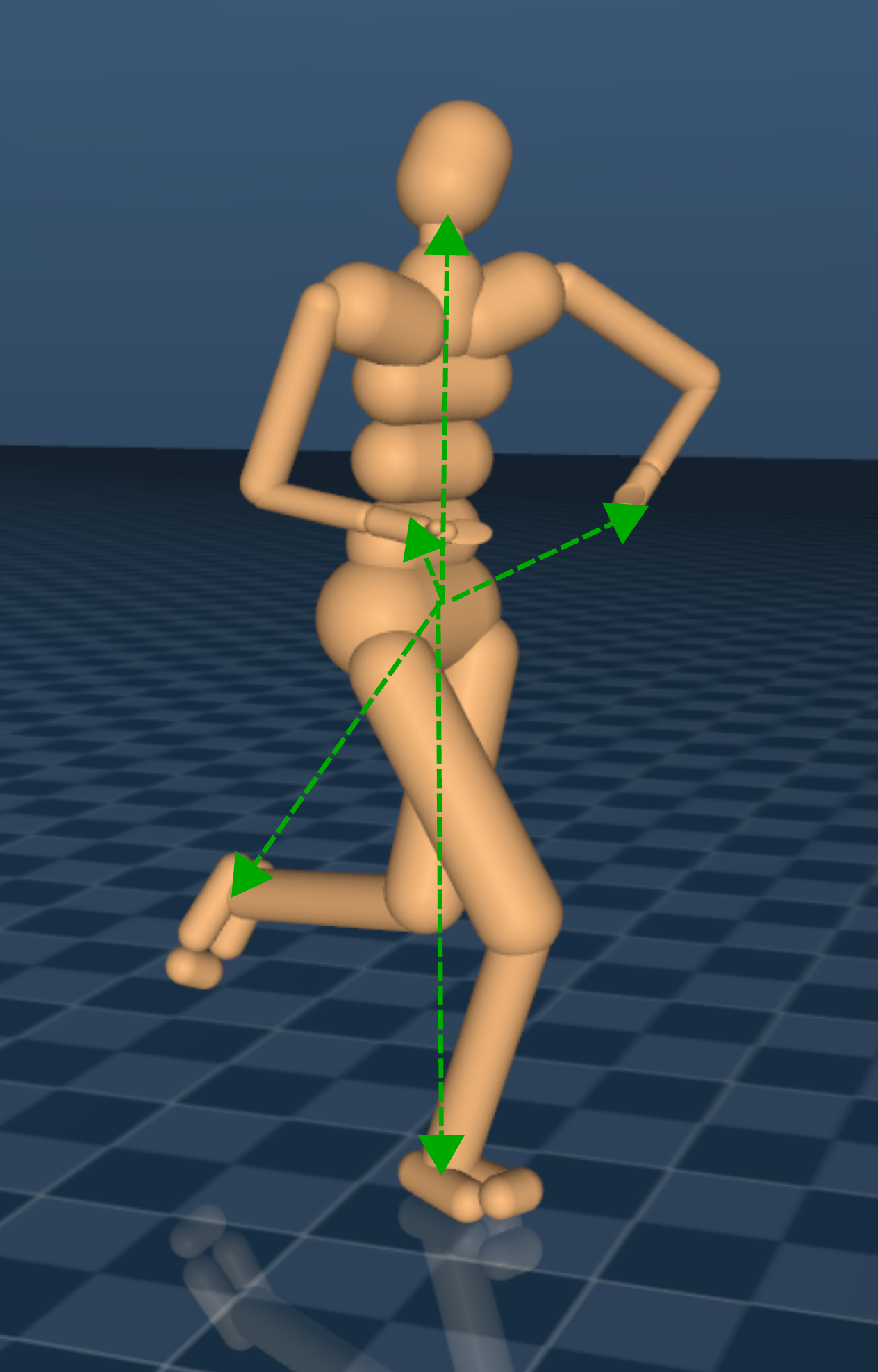}
\end{subfigure}
        \begin{tabular}{c}
            \begin{subfigure}[t]{0.77\textwidth}
                \includegraphics[width=1\textwidth]{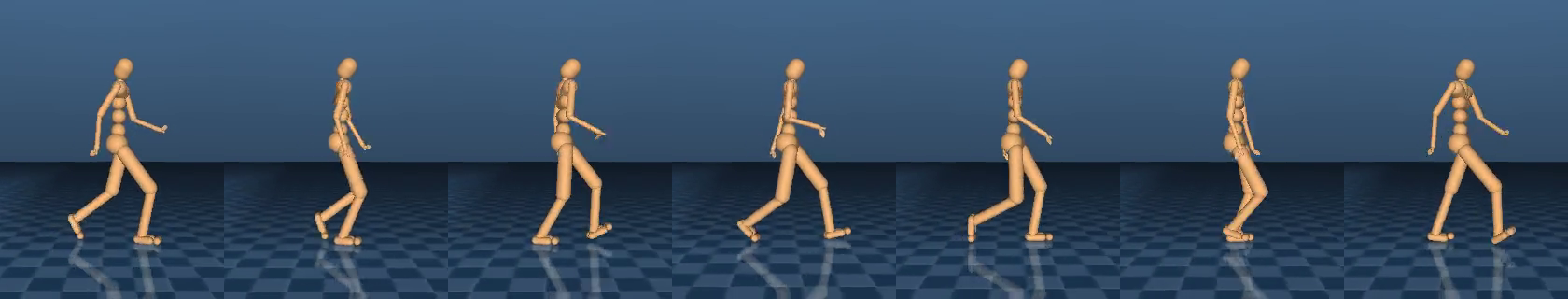}
            \end{subfigure}\\
            \begin{subfigure}[t]{0.77\textwidth}
                \includegraphics[width=1\textwidth]{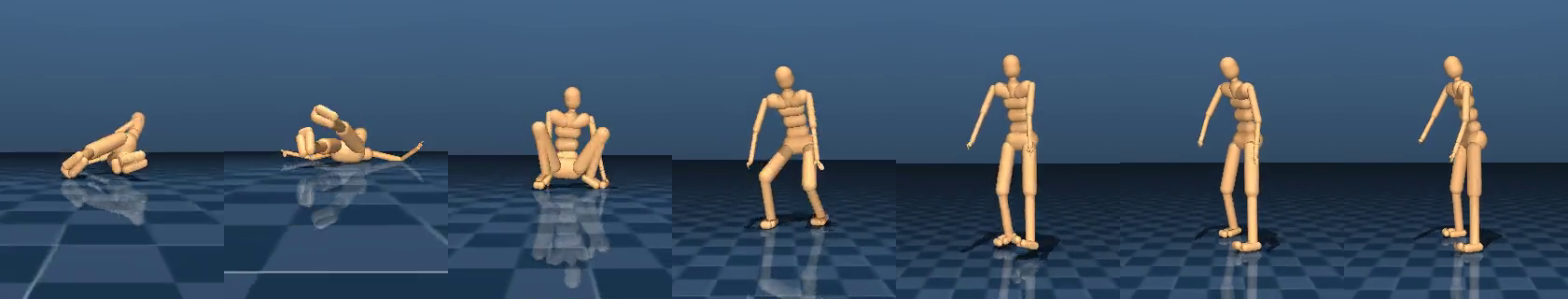}
            \end{subfigure}
        \end{tabular}\\
    \end{tabular}
    \caption{(Left) Example frame of humanoid body with illustration of end-effector features corresponding to 3D vectors from the root to the feet, hands, and head.  (Right) Equispaced frames from an execution of policies trained by imitation from motion capture to perform a walking behavior (top) and getting up behavior (bottom).}
    \label{complexHumanoid}
\end{figure}

We consequently removed all non-egocentric features provided to either the policy or discriminator. Additionally, we provided both the policy and discriminator with built-in mujoco sensors corresponding to a ``velocimeter'' that computes velocities axis-aligned to the root orientation frame, a ``gyroscope" that provides egocentric rotational velocity, an ``accelerometer" that indicates uprightness, as well as custom 3D vectors from the root (pelvis) to the feet (x2), hands (x2), and head (see Figure \ref{complexHumanoid}, left panel for schematic depiction of end-effector features). 
Analogous to learning from end-effector features in the arm case, we only exposed these indirect measurements (velocimeter, gyros, and end-effector vectors) to the discriminator. With these features, the apparent quality of the imitation behavior dramatically improved (Figure \ref{complexHumanoid}, top-right, or \href{https://www.youtube.com/watch?v=qynRl_NoRCY}{{\color{blue}video}}).  

\textbf{Getting up from the ground:} To examine a qualitatively distinct behavior, we attempted to learn a policy to get up from the ground in a humanlike way. For this task, we added the imitation-reward from the discriminator with a small reward for having the head at standing height. 
In this case we used a mixture of clips for getting up from a diversity of starting states (see supplement). A moderately robust policy was learned that gets up from a variety of starting states and stands (see Figure \ref{complexHumanoid}, right-bottom, or \href{https://www.youtube.com/watch?v=Fx2bPFmXkzc}{{\color{blue}video}}). While many seeds learned to robustly get up, not all seeds learned to remain standing after getting up (instead, looping the standing up behavior and then falling).

\textbf{Stylized movement from scratch, using single demonstrations:} To further explore single behavior acquisition from limited demonstration data, we attempted to imitate non-standard gait styles using single clips.  We attempted this for small number of clips without a hyperparameter search (for the simple walk, various but not all hyperparameters performed reasonably, and we selected a single successful hyperparameter setting to try here) and while imitation from single clips did not yield perfect behavior and did not work for every clip we tried, some resulting policies were robust and captured the essence of the style (e.g. \href{https://www.youtube.com/watch?v=gJ5H5MvnOig}{{\color{blue}drunk-style policy}}, \href{https://www.youtube.com/watch?v=kA3FJNX37qA}{{\color{blue}chicken-style policy}}; videos include user-interactive application of small exogenous forces to guide/redirect the movement).  We view it as quite non-trivial that it is possible to train a policy to generate a moderately robust behavior from a single clip (roughly 10s of a person with a different body performing that behavior).

In addition to the above behaviors, we trained policy sub-skills for running and turning (\href{https://www.youtube.com/watch?v=ZNOVYLjWrto}{{\color{blue}video}}).

\subsection{Contextual modulation and task-based control}

\begin{figure}[t!]
  \centering
  \includegraphics[width=0.335\linewidth]{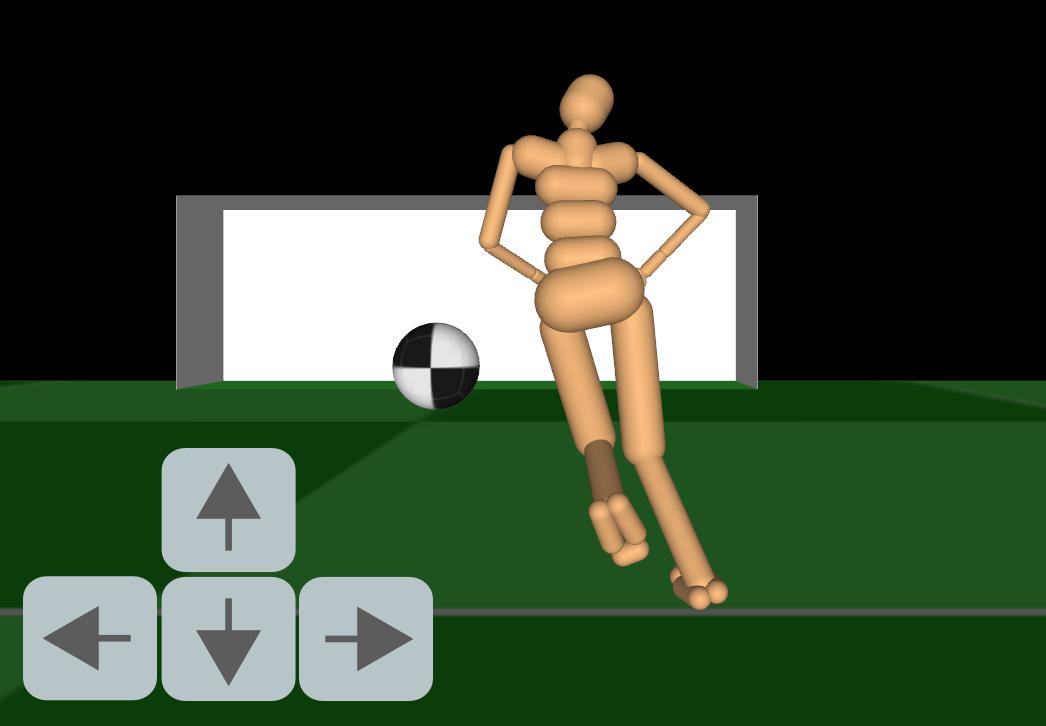}
  \includegraphics[width=0.655\linewidth]{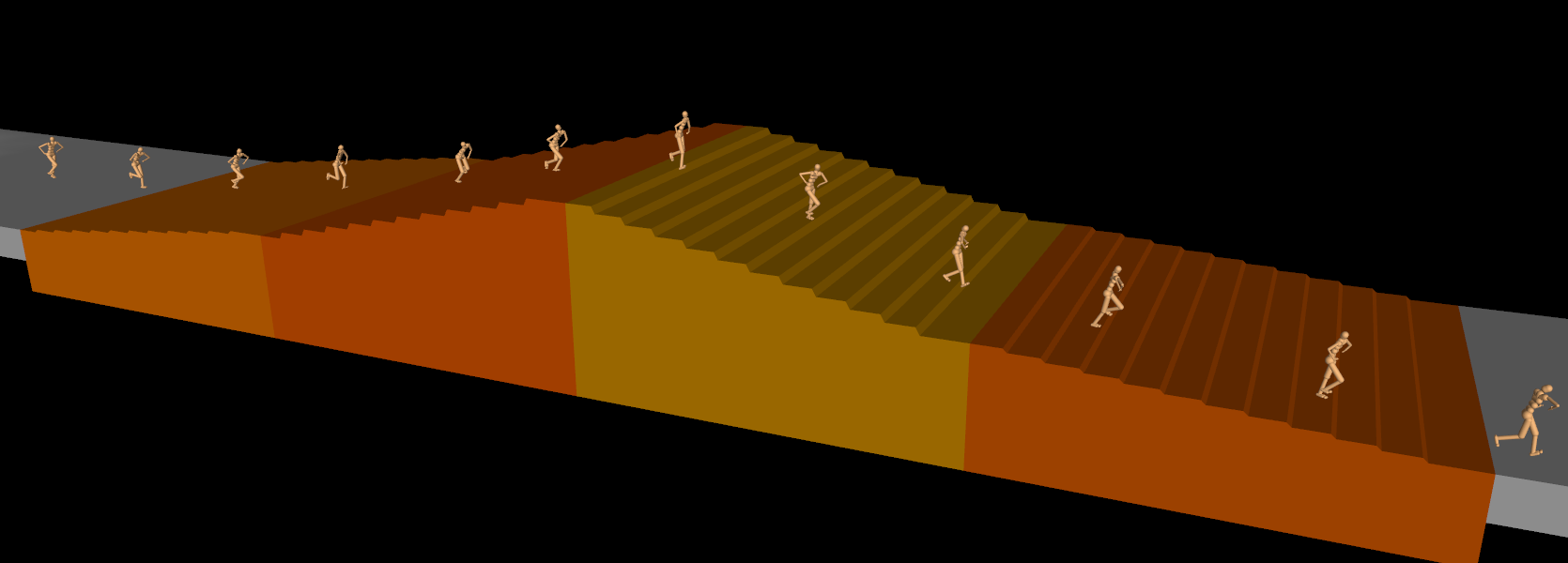}
  \includegraphics[width=1\linewidth]{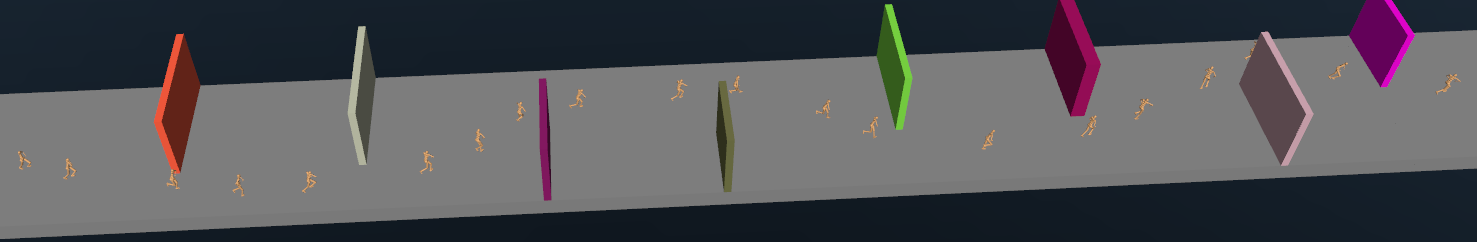}
  \caption{Depictions of higher-level tasks. (Top left) Soccer gameplay using the keyboard. (Top right) Image depicting the ascend and descent of the humanoid over stairs, after policy fine-tuning. (Bottom) Image depicting the movement of a humanoid along a course, controlled by a higher-level control which determines modulates a pre-trained lower-level controller which could run and turn.}
  \label{composition}
\end{figure}

Up to this point, all policies we have discussed are trained to perform one skill. 
To build systems capable of flexible behavior, we want to transition among sub-skills.  
We found that switching mid-trajectory between separately-trained, single-skill policies did not lead to robust transitioning as the intermediate states were outside the state distributions of one or the other policy.

We found that adversarial imitation training using one context per behavior, as described in Figure \ref{diagram}, encourages the policy to find efficient transitions between behaviors. 
Accordingly, we found it possible to train a single low-level controller capable of switching among running straight, veering left, and veering right based on an externally set one-of-three context (from a total of only 286 time-resampled frames; <10s of data). During training, the context was random and switched every 5s, but at run-time, it is possible to modulate it interactively to control the character. 
This multi-skill, low-level controller could be controlled either by keyboard-determined modulation of the context, or, as we will show, by a higher-level controller.

As a first example, we restored the low-level controller in a new environment corresponding to a half of a soccer field and modulate the low-level controller by keyboard-control to score a goal (Figure \ref{composition}, or \href{https://www.youtube.com/watch?v=UrXaSB7iRhE}{{\color{blue}video}}).  

Additionally, we considered whether it is possible to refine previously learned skills. We exposed the existing policy and body to a staircase track and fine-tuned the running policy to surmount stairs by RL with a simple reward for forward velocity
(Figure \ref{composition}, or
\href{https://www.youtube.com/watch?v=HPwPAF06QmY}{{\color{blue}video}}). Without tuning, the running policy falls quickly on the stairs, but the refined policy is able to ascend and descend stairs of two discrete slopes, while retaining its ability to locomote on flat ground.

Finally, we considered modulation of the controller by a higher-level controller. 
The higher level controller could be trained by a variety of methods and could interact with the policy in a variety of ways. 
Here, we used a simple two-layer neural network as the higher-level controller and trained it to use a local, top-down depth camera to send context signals to the lower-level controller, resulting in autonomous navigation. Its reward corresponded to forward movement along a linear (3D) track (see Figure \ref{composition} or \href{https://www.youtube.com/watch?v=3iiJgo01eoE}{{\color{blue}video}}: the probe trial is a representative rollout on a novel procedurally generated instance of the course). See supplemental information for additional details.

\section{Relationships with related work}

\textbf{Generative sequence modeling:} There has been a great deal of research into sequence models of plausible movements, not in the context of physical control.  For example, modeling the kinematic trajectories of a motion capture data in the absence of physics has been actively studied in the neural network literature \cite{taylor2007modeling,sussillo2009generating,jaeger2014conceptors}. 

\textbf{Physical control:} 
In the field of graphics-oriented character animation, much attention has been paid to building realistic movement behaviors, though the results often require significant domain expertise.
Representative methods for producing terrestrial locomotion have made use of specialized gait controllers (for a quadruped \cite{coros2011locomotion}; for a biped \cite{mordatch2010robust}) or systems explicitly producing foot placements \cite{agrawal2016task}. Curated motion capture has been employed, for example, in beautiful work to simulate pigeon flight \cite{ju2013data} as well as in the context of high performance open-loop trajectory planning for humanoid behaviors \cite{liu2010sampling}. 
These tools have supported amazing demonstrations of sequenced sub-skills in obstacle courses \cite{liu2012terrain}; however, the sub-skills are each individually highly-engineered, albeit quite elegantly.
Trajectory optimization (or planning) without motion capture can also yield impressive results (e.g. \cite{al2013trajectory,hamalainen2014online}). 
Furthermore, trajectories optimized using model-based approaches can be distilled into a neural network policy \cite{mordatch2015interactive}. 

Deep RL methods, by contrast, provide hope that many of these results can be replicated and perhaps improved upon in the near future with substantially less domain knowledge.
Exciting recent work has used deep RL for 2D locomotion of creatures \cite{peng2016terrain} and very recently for 3D humanoid \cite{2017-TOG-deepLoco}.  
Other recent work successfully makes use of  a specialized neural network architecture for motion capture sequence modeling to enable robust humanoid behavior \cite{holden2017phase}. 
While these have used comparatively less domain knowledge than the other approaches in animation, the priority in these papers has been to produce aesthetically pleasing motions over constructing generic tools. 
Our work makes fewer assumptions about how policies should track the motion capture data, uses a more general learning approach, and builds in less structure in controllers and reward functions.  

\section{Discussion}

Looking beyond the scope of engineering skilled motor behaviors for humanoids, we think there is a broader issue to consider in the design of artificial agents.
To communicate complex behaviors to agents, it is often most straightforward to demonstrate them.
In contrast, it is extremely difficult to formalize different behaviors with simple reward functions. 

Therefore, a core motivation to use imitation learning is that we lack good objective functions to describe complex behaviors. This necessarily presents an obstacle when developing and assessing algorithms as well as when monitoring convergence (not dissimilar from the difficulty of assessing generated samples from GANs). At present, we must rely on human judgment of the quality of the behaviors we have produced.

So far, we have only demonstrated learning on a somewhat restricted, albeit diverse, set of behaviors and have only demonstrated limited reuse.  
Accordingly, subsequent steps in this research program will include scaling to a much wider behavioral repertoire and leveraging the learned sub-skills for even more challenging tasks.

\subsubsection*{Acknowledgments}

We thank Matt Botvinick and Scott Reed for useful discussion as well as others at DeepMind for support.
The data used in this project was obtained from mocap.cs.cmu.edu.
The database was created with funding from NSF EIA-0196217.

\setlength{\bibsep}{1pt plus 0.3ex}

\newpage
\appendix
\renewcommand{\thefigure}{A\arabic{figure}}
\setcounter{figure}{0}

\section*{Supplemental information:}

\thispagestyle{plain}


\section{Additional motion capture details}

The CMU database provides Acclaim Skeleton Files (ASF) and Acclaim Motion Capture (AMC), corresponding respectively to the segment-lengths of the bodies of the subjects and the joint-angle poses per frame.  The complex humanoid model used in this work was programmatically generated from subject 8 from the CMU database by taking the body segment lengths available in the ASF and asserting reasonable values for other body proportions as well as actuator strengths. 

The demonstration data comes from loading the AMC files, resampling from native motion capture rate of 120Hz down to 30ms timesteps by spline interpolation, setting the pose of the body (i.e. static configuration of joint angles) to a frame from the AMC file, and logging features computed from the body state as observations (see results for more detail). 

\section{Additional experimental details}

When training with TRPO for RL, we performed online z-filtering of inputs to the networks as well as rewards. When performing imitation learning, we did not filter the rewards, but we clipped them (max value of 10).  Advantages were normalized per iteration.  Action variance sigma (parameterized by log-sigma) was generally initialized to be between $e^{-1}\approx.37$ and $1$. 

\subsection*{Experiments using walker:}
Observations available to the policy consist of state values for vertical position of the body as well as all joint angles, angular velocities of the joints, and horizontal and vertical velocity -- these same observations are available to the discriminator for adversarial imitation.
We logged trajectories from the RL-trained policy and trained imitation policies to match the behavior.  All walker experiments presented here used 30,000 samples per iteration and adversarial imitation from RL demonstrations use 100 demonstration episodes.  We performed a small sweep over discriminator hyperparameters -- learning rates: [1e-5, 1e-4, 1e-3]; number of updates to the discriminator per iteration: [1, 5, 10, 20].  The policy architecture consisted of a multi-layer neural network with two hidden layers (200, 100) with hyperbolic tangent nonlinearities.

\begin{figure}[h!]
  \centering
  \includegraphics[width=1\linewidth]{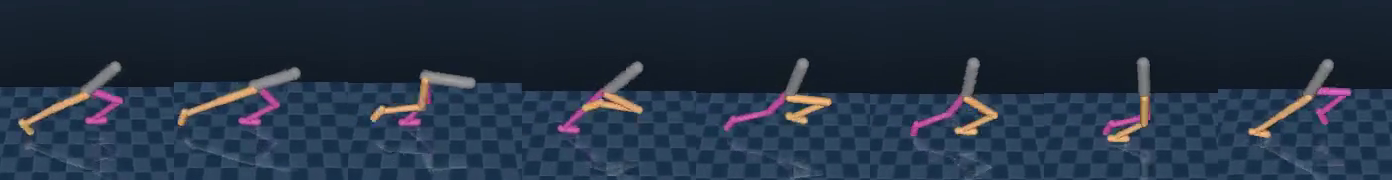}
  \caption{Equispaced frames of a baseline RL-trained walker running.}
  \label{walker_viz}
\end{figure}

\subsection*{Experiments using arms:}
Observations available to the policy consist of the sines and cosines of the joint angles as well as the angular velocities of the joints and finally a vector from the arm end-effector to the center of the target (note that the goal position also works).   
All arm experiments presented here used 20,000 samples per iteration and adversarial imitation from RL demonstrations use 100 demonstration episodes.  Again, the discriminator hyperparameters were swept -- learning rates: [1e-4, 1e-3, 1e-2] ; number of updates to the discriminator per iteration: [1, 5, 10, 20].  The policy architecture consisted of a multi-layer neural network with two hidden layers (100, 60) with hyperbolic tangent nonlinearities.

\begin{figure}[h!]
  \centering
  \includegraphics[width=1\linewidth]{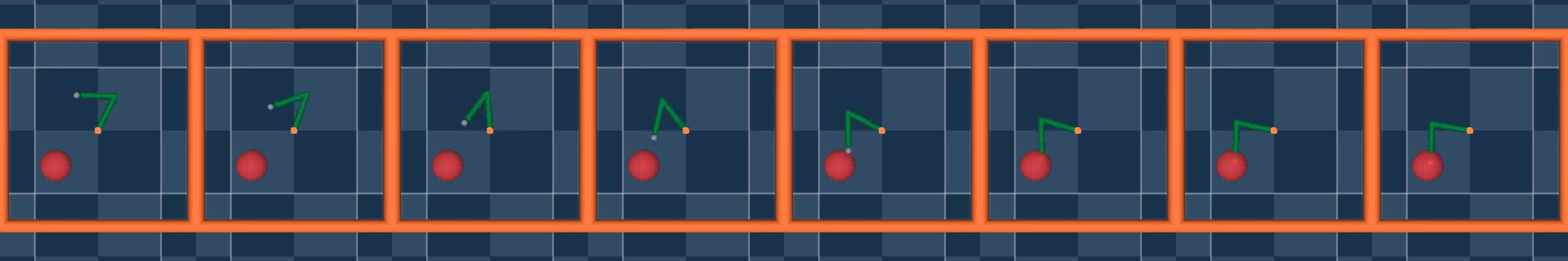}
  \includegraphics[width=1\linewidth]{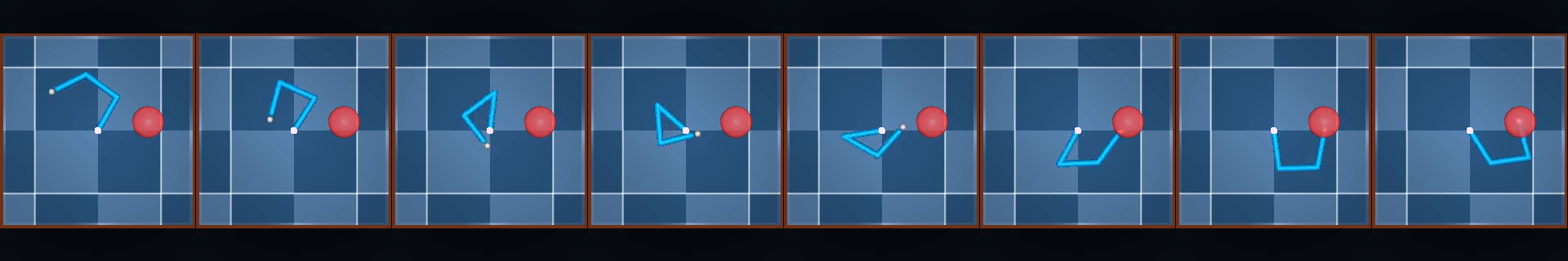}
  \caption{Equispaced frames of a baseline RL-trained (top) two-link arm, (bottom) three-link arm.}
  \label{reacher_viz}
\end{figure}

\subsection*{Experiments using complex humanoid:}

The environments which make use of this humanoid body, have a simulation timestep of 5ms and a control timestep of 30ms.
We also note that actions from the neural network policies are bounded [-1,1], and joint-angle torque-generators have gains associated with each action dimension (e.g. legs have high gain and toes have a small gain).

For RL, the environment is constituted of a plane on which the humanoid body is placed at the origin and the task is for the body to move forward.  For RL training of locomotion, the reward is the instantaneous forward velocity, with a small control cost.  Episodes are 10 seconds long and falling causes an early episode termination -- this implicitly affects the reward and also biases the states visited to those that may be expected to be more reasonable.

When using a forward reward, the reward consisted of a baseline value simply for staying alive (i.e. not falling and triggering an early termination), a small control cost on the squared actions, a term encouraging staying in the middle of the track (i.e. not veering horizontally), and a term which either promoted running forward as fast as possible or penalized the absolute difference from a target velocity.

We note that while training policies for the simpler environments by RL took well under 500 iterations of TRPO, the complex humanoid took between 1000 and 1500 iterations to fully saturate.  
All humanoid experiments presented here used 100,000 samples per iteration.
Imitation in the simpler environments seemed to take roughly the same amount of time as RL, so when imitating from motion capture, we ran experiments for up to 3000 iterations, assuming this would be conservatively high.  Resulting policies often appeared (visually) to saturate in fewer than this maximum number of iterations.

For imitation tasks, we swept only over discriminator learning rate, number of discriminator updates per iteration, and for humanoid locomotion, we used a generalized advantage estimation discount coefficient 
[1].  
The policy architecture consisted of a multi-layer neural network with three hidden layers (300, 200, 100) with hyperbolic tangent nonlinearities.  For imitation learning from walking behaviors, we also used early termination of episodes when the body fell.  Specific motion capture data used per behavior is detailed in a table below:

\begin{table}[H]
  \caption*{Motion capture clips}
  \label{sample-table}
  \centering
  \begin{tabular}{lll}
    \toprule
    Type     & Subject     & clips \\
    \midrule
    simple walk & 8  & 1-11     \\
    get up*     & 77 & 16, 17, 18      \\
         & 140 & 1, 2, 3, 4, 8, 9      \\
    drunk     & 137       & 16  \\
    chicken     & 137       & 8  \\
    running and turning & 16       & 45, 46, 51, 52, 53, 54, 56  \\
    \bottomrule
  \end{tabular}
\end{table}

*Note that the get-up behavior pooled data from two subjects.

Walls navigation experiments were performed using an approximate version of TRPO (proximal policy optimization, or PPO [2]) which is amenable to parallelization (the results should not be expected to differ meaningfully than those obtainable by TRPO).  
The higher level controller receives observations corresponding to a top down depth map of a rectangular region up to 15m ahead of the body and horizontally corresponding to 80\% of the width of the track (8m).  This distance corresponds to it seeing roughly two walls ahead at a time.  Anecdotally, when navigating by keyboard, it also makes sense to look more than one wall ahead when possible to prepare for turns -- the information available to the agent was designed with this consideration in mind.  The state available to the higher level controller also includes egocentric heading and lateral position but does not include position along the length of the track.  The higher-level controller consisted only of a generic 2-layer MLP (i.e. no special NN policy architecture was used to process the depth map features). 






{\small 
\begin{thebibliography}{22}
\providecommand{\natexlab}[1]{#1}
\providecommand{\url}[1]{\texttt{#1}}
\expandafter\ifx\csname urlstyle\endcsname\relax
  \providecommand{\doi}[1]{doi: #1}\else
  \providecommand{\doi}{doi: \begingroup \urlstyle{rm}\Url}\fi

\bibitem[Rosheim(2006)]{rosheim2006leonardo}
Mark Rosheim.
\newblock \emph{Leonardo's Lost Robots}.
\newblock Springer Science \& Business Media, 2006.

\bibitem[Ho and Ermon(2016)]{ho2016generative}
Jonathan Ho and Stefano Ermon.
\newblock Generative adversarial imitation learning.
\newblock In \emph{Advances in Neural Information Processing Systems}, pages
  4565--4573, 2016.

\bibitem[Goodfellow et~al.(2014)Goodfellow, Pouget-Abadie, Mirza, Xu,
  Warde-Farley, Ozair, Courville, and Bengio]{goodfellow2014generative}
Ian Goodfellow, Jean Pouget-Abadie, Mehdi Mirza, Bing Xu, David Warde-Farley,
  Sherjil Ozair, Aaron Courville, and Yoshua Bengio.
\newblock Generative adversarial nets.
\newblock In \emph{Advances in neural information processing systems}, pages
  2672--2680, 2014.

\bibitem[Schulman et~al.(2015{\natexlab{a}})Schulman, Levine, Abbeel, Jordan,
  and Moritz]{schulman2015trust}
John Schulman, Sergey Levine, Pieter Abbeel, Michael~I Jordan, and Philipp
  Moritz.
\newblock Trust region policy optimization.
\newblock In \emph{ICML}, pages 1889--1897, 2015{\natexlab{a}}.

\bibitem[Todorov et~al.(2012)Todorov, Erez, and Tassa]{todorov2012mujoco}
Emanuel Todorov, Tom Erez, and Yuval Tassa.
\newblock Mujoco: A physics engine for model-based control.
\newblock In \emph{Intelligent Robots and Systems (IROS), 2012 IEEE/RSJ
  International Conference on}, pages 5026--5033. IEEE, 2012.

\bibitem[Heess et~al.(2016)Heess, Wayne, Tassa, Lillicrap, Riedmiller, and
  Silver]{heess2016learning}
Nicolas Heess, Greg Wayne, Yuval Tassa, Timothy Lillicrap, Martin Riedmiller,
  and David Silver.
\newblock Learning and transfer of modulated locomotor controllers.
\newblock \emph{arXiv preprint arXiv:1610.05182}, 2016.

\bibitem[Schulman et~al.(2015{\natexlab{b}})Schulman, Moritz, Levine, Jordan,
  and Abbeel]{schulman2015high}
John Schulman, Philipp Moritz, Sergey Levine, Michael Jordan, and Pieter
  Abbeel.
\newblock High-dimensional continuous control using generalized advantage
  estimation.
\newblock \emph{arXiv preprint arXiv:1506.02438}, 2015{\natexlab{b}}.

\bibitem[Taylor et~al.(2007)Taylor, Hinton, and Roweis]{taylor2007modeling}
Graham~W Taylor, Geoffrey~E Hinton, and Sam~T Roweis.
\newblock Modeling human motion using binary latent variables.
\newblock \emph{Advances in neural information processing systems},
  19:\penalty0 1345, 2007.

\bibitem[Sussillo and Abbott(2009)]{sussillo2009generating}
David Sussillo and Larry~F Abbott.
\newblock Generating coherent patterns of activity from chaotic neural
  networks.
\newblock \emph{Neuron}, 63\penalty0 (4):\penalty0 544--557, 2009.

\bibitem[Jaeger(2014)]{jaeger2014conceptors}
Herbert Jaeger.
\newblock Conceptors: an easy introduction.
\newblock \emph{arXiv preprint arXiv:1406.2671}, 2014.

\bibitem[Coros et~al.(2011)Coros, Karpathy, Jones, Reveret, and Van
  De~Panne]{coros2011locomotion}
Stelian Coros, Andrej Karpathy, Ben Jones, Lionel Reveret, and Michiel Van
  De~Panne.
\newblock Locomotion skills for simulated quadrupeds.
\newblock \emph{ACM Transactions on Graphics (TOG)}, 30\penalty0 (4):\penalty0
  59, 2011.

\bibitem[Mordatch et~al.(2010)Mordatch, De~Lasa, and
  Hertzmann]{mordatch2010robust}
Igor Mordatch, Martin De~Lasa, and Aaron Hertzmann.
\newblock Robust physics-based locomotion using low-dimensional planning.
\newblock \emph{ACM Transactions on Graphics (TOG)}, 29\penalty0 (4):\penalty0
  71, 2010.

\bibitem[Agrawal and van~de Panne(2016)]{agrawal2016task}
Shailen Agrawal and Michiel van~de Panne.
\newblock Task-based locomotion.
\newblock \emph{ACM Transactions on Graphics (TOG)}, 35\penalty0 (4):\penalty0
  82, 2016.

\bibitem[Ju et~al.(2013)Ju, Won, Lee, Choi, Noh, and Choi]{ju2013data}
Eunjung Ju, Jungdam Won, Jehee Lee, Byungkuk Choi, Junyong Noh, and Min~Gyu
  Choi.
\newblock Data-driven control of flapping flight.
\newblock \emph{ACM Transactions on Graphics (TOG)}, 32\penalty0 (5):\penalty0
  151, 2013.

\bibitem[Liu et~al.(2010)Liu, Yin, van~de Panne, Shao, and Xu]{liu2010sampling}
Libin Liu, KangKang Yin, Michiel van~de Panne, Tianjia Shao, and Weiwei Xu.
\newblock Sampling-based contact-rich motion control.
\newblock \emph{ACM Transactions on Graphics (TOG)}, 29\penalty0 (4):\penalty0
  128, 2010.

\bibitem[Liu et~al.(2012)Liu, Yin, van~de Panne, and Guo]{liu2012terrain}
Libin Liu, KangKang Yin, Michiel van~de Panne, and Baining Guo.
\newblock Terrain runner: control, parameterization, composition, and planning
  for highly dynamic motions.
\newblock \emph{ACM Trans. Graph.}, 31\penalty0 (6):\penalty0 154, 2012.

\bibitem[Al~Borno et~al.(2013)Al~Borno, De~Lasa, and
  Hertzmann]{al2013trajectory}
Mazen Al~Borno, Martin De~Lasa, and Aaron Hertzmann.
\newblock Trajectory optimization for full-body movements with complex
  contacts.
\newblock \emph{IEEE transactions on visualization and computer graphics},
  19\penalty0 (8):\penalty0 1405--1414, 2013.

\bibitem[H{\"a}m{\"a}l{\"a}inen et~al.(2014)H{\"a}m{\"a}l{\"a}inen, Eriksson,
  Tanskanen, Kyrki, and Lehtinen]{hamalainen2014online}
Perttu H{\"a}m{\"a}l{\"a}inen, Sebastian Eriksson, Esa Tanskanen, Ville Kyrki,
  and Jaakko Lehtinen.
\newblock Online motion synthesis using sequential monte carlo.
\newblock \emph{ACM Transactions on Graphics (TOG)}, 33\penalty0 (4):\penalty0
  51, 2014.

\bibitem[Mordatch et~al.(2015)Mordatch, Lowrey, Andrew, Popovic, and
  Todorov]{mordatch2015interactive}
Igor Mordatch, Kendall Lowrey, Galen Andrew, Zoran Popovic, and Emanuel~V
  Todorov.
\newblock Interactive control of diverse complex characters with neural
  networks.
\newblock In \emph{Advances in Neural Information Processing Systems}, pages
  3132--3140, 2015.

\bibitem[Peng et~al.(2016)Peng, Berseth, and van~de Panne]{peng2016terrain}
Xue~Bin Peng, Glen Berseth, and Michiel van~de Panne.
\newblock Terrain-adaptive locomotion skills using deep reinforcement learning.
\newblock \emph{ACM Transactions on Graphics (TOG)}, 35\penalty0 (4):\penalty0
  81, 2016.

\bibitem[Peng et~al.(2017)Peng, Berseth, Yin, and van~de
  Panne]{2017-TOG-deepLoco}
Xue~Bin Peng, Glen Berseth, KangKang Yin, and Michiel van~de Panne.
\newblock Deeploco: Dynamic locomotion skills using hierarchical deep
  reinforcement learning.
\newblock \emph{ACM Transactions on Graphics (Proc. SIGGRAPH 2017)},
  36\penalty0 (4), 2017.

\bibitem[Holden et~al.(2017)Holden, Komura, and Saito]{holden2017phase}
Daniel Holden, Taku Komura, and Jun Saito.
\newblock Phase-functioned neural networks for character control.
\newblock \emph{ACM Transactions on Graphics (Proc. SIGGRAPH 2017)}, 2017.

\end{thebibliography}

}

\section*{References}

\vspace{-.5cm}
\medskip

\small

[1] John Schulman, Philipp Moritz, Sergey Levine, Michael Jordan, and Pieter Abbeel.  High-dimensional continuous control using generalized advantage estimation.arXiv preprint arXiv:1506.02438, 2015.

[2] Pieter Abbeel and John Schulman. Deep reinforcement learning through policy optimization. Tutorial at Neural Information Processing Systems, 2016.  URL https://nips.cc/Conferences/2016/Schedule?showEvent=6198.




\end{document}